\documentclass[10pt,twocolumn,letterpaper]{article}

\usepackage[pagenumbers]{cvpr} 

\usepackage{multirow}
\usepackage{amsmath}
\usepackage{mathtools} 
\usepackage{algorithm2e}

\definecolor{cvprblue}{rgb}{0.21,0.49,0.74}
\usepackage[pagebackref,breaklinks,colorlinks,citecolor=cvprblue]{hyperref}

\newcommand*{\huzhang}[1]{\textcolor{orange}{[huzhang: #1]}}

\definecolor{teal}{rgb}{0,0.6,0.6}

\def\paperID{6498} 
\def\confName{CVPR}
\def\confYear{2024}

\title{Exploring Transferability for Randomized Smoothing}

\author{%
Kai Qiu \qquad
Huishuai Zhang \qquad
Zhirong Wu \qquad
Stephen Lin \\
Microsoft Research Asia\\
{\tt\small \{kai.qiu, huishuai.zhang, wu.zhirong, stevelin\}@microsoft.com}
}

\begin{document}
\maketitle
\begin{abstract}

Training foundation models on extensive datasets and then finetuning them on specific tasks has emerged as the mainstream approach in artificial intelligence. 
	However, the model robustness, which is a critical aspect for safety, is often optimized for each specific task rather than at the pretraining stage.
	In this paper, we propose a method for pretraining certifiably robust models that can be readily finetuned for adaptation to a particular task. A key challenge is dealing with the compromise between semantic learning and robustness. We address this with a simple yet highly effective strategy based on significantly broadening the pretraining data distribution, which is shown to greatly benefit finetuning for downstream tasks. Through pretraining on a mixture of clean and various noisy images, we find that surprisingly strong certified accuracy can be achieved even when finetuning on only clean images. Furthermore, this strategy requires just a single model to deal with various noise levels, thus substantially reducing computational costs in relation to previous works that employ multiple models. Despite using just one model, our method can still yield results that are on par with, or even superior to, existing multi-model methods.
\end{abstract}    
\section{Introduction}
\label{sec:intro}
Refining pre-trained foundation models for task-specific applications has become the leading strategy in AI. 
These foundation models acquire rich representations from a broad data distribution during the pre-training phase.
Consequently, a slight amount of fine-tuning is sufficient for these foundation models to competently adapt to various downstream tasks.
This paradigm has become prevalent in various domains, including but not limited to natural language processing \cite{kenton2019bert,radford2019language}, computer vision \cite{chen2020simple,radford2021learning, alayrac2022flamingo, he2022masked} and multimedia \cite{li2023multimodal}.

\begin{table}[]
  \centering
  \caption{Transferability of semantics and robustness. Mixed noise pre-training allows the model to learn from a broad data distribution, achieving strong certified robustness even when fine-tuned only on clean images. Even just one epoch of fine-tuning can yield strong natural and certified accuracy. We perform experiments under two configurations: fine-tuning the full network (top half of the table) and only fine-tuning the last FC layer with fixed features (bottom half of the table). We pre-train on ImageNet and fine-tune on CIFAR10.
   ``Clean" refers to training only with clean images. $\sigma=0.25$ means training with noise of standard deviation 0.25 added to the clean image. ``Mixed Noise" indicates training by mixing clean images with Gaussian noise of various $\sigma$. The $\star$ represents fine-tuning for just one epoch. The top two results in each metric are indicated in bold.}
  \label{tab:sum}
  \begin{tabular}{cccc}
    \hline

  \hline
  \multicolumn{4}{c}{Full-Network}                                                                                               \\ \hline
  Pre-train           & Fine-tune           & Clean Acc. & \begin{tabular}[c]{@{}c@{}}Certified Acc. \\  at $\varepsilon$ = 0.25 (\%)\end{tabular} \\ \hline
  Clean              & Clean              & \textbf{97.8}       & 10.1                                                                    \\
  Clean              & $\sigma =$ 0.25 & 77.6       & 69.4                                                                    \\
  $\sigma =$ 0.25 & Clean              & 94.4       & 21.6                                                                    \\
  $\sigma =$ 0.25 & $\sigma =$ 0.25 & 83.1       & \textbf{72.6}                                                                    \\ \hline
  Mixed Noise        & Clean $ \star $              & 96.6       & 65.9                                                                    \\ 
  Mixed Noise        & Clean              & \textbf{98.0}       & \textbf{71.4}                                                                    \\ \hline
  \hline
  \multicolumn{4}{c}{Fixed-Feature}                                                                                              \\ \hline
  Pre-train           & Fine-tune           & Clean Acc. & \begin{tabular}[c]{@{}c@{}}Certified Acc. \\  at $\varepsilon$ = 0.25 (\%)\end{tabular} \\ \hline
  Clean              & Clean              & 94.3       & 10.1                                                                    \\
  Clean              & $\sigma =$ 0.25 &   23.8     &   16.5                                                                  \\
  $\sigma =$ 0.25 & Clean       &  84.8      &    65.9                                                                 \\
  $\sigma =$ 0.25 & $\sigma =$ 0.25 & 77.9       & \textbf{72.6}                                                                    \\ \hline
  Mixed Noise        & Clean $ \star $              &  \textbf{94.7}      &  65.3                                                                   \\ 
  Mixed Noise        & Clean              &  \textbf{96}       & \textbf{67.3}                                                                    \\ \hline

  \hline
  \end{tabular}
\end{table}

Robustness of models to adversarial attacks has become increasingly important with the proliferation of AI to many real-world applications, and much research has focused on making models robust by including numerous adversarial examples to the training data \cite{madry2018towards,cohen2019certified}. However, a substantial enlargement of the training data at the finetuning stage would counter the aim of finetuning to reduce training costs. This raises the question of whether robustness can be instilled in the pre-trained model and transferred to downstream tasks in a manner similar to semantic knowledge. This problem of robustness transfer has recently been examined in the context of adversarial training~\cite{shafahi2019adversarially, salman2020adversarially}, where a robustness objective is also optimized during pre-training. Such an approach, though, does not offer {\em certifiable robustness} that ensures reliable outcomes for inputs with a certain degree of perturbation. Furthermore, their gains in robustness come at the expense of natural accuracy on clean images.

In this paper, our goal is to pre-train a model that can be readily fine-tuned for adaption to a specific task while maintaining high accuracy on clean images and strong certified robustness.
Among methods for establishing certified robustness, the dominant approach is randomized smoothing, which trains models on random Gaussian noise and utilizes a smoothed classifier to ensure robustness in a certain perturbation radius.
However, a key challenge of both randomized smoothing and previous robustness transfer techniques is in dealing with the trade-offs between semantic learning and robustness.
Our findings indicate the importance of expanding the data distribution in the pre-training stage to address this issue.
We show how a broad data distribution enables the model to learn both rich semantic features and robustness simultaneously.
Moreover, we find that such a model does not require any additional finetuning on noisy images to preserve robustness and adapt to downstream tasks.
To expand the data distribution, we present a simple, yet incredibly powerful strategy.
In particular, we pre-train on a mixture of clean and noisy images of various noise radius.

Surprisingly, we find that such a mixed pre-trained model, even when fine-tuned only on clean images of the downstream tasks, can achieve a strong level of certified robustness (see Table \ref{tab:sum}).
This indicates that the method has successfully transferred not only the semantic features but also robustness to the downstream tasks.
In a more extreme scenario, even with only one epoch of fine-tuning, our method can still yield good results.
This allows for a fine-tuning process that is remarkably straightforward and efficient.
Existing works on randomized smoothing often sacrifice natural accuracy in order to achieve robust accuracy.
In contrast, our method does not compromise on clean data accuracy, yielding strong transfer performance on semantics and robustness simultaneously.

Another advantage of our strategy is that it only requires training a single model to handle various levels of noise, instead of training multiple models.
Existing research on randomized smoothing involves training a separate model for each particular noise radius.
Such trained models can only handle specific noise and struggle when faced with attacks of different magnitudes.
Also, training multiple models complicates the model's deployment process.
In contrast, our approach needs only one model to handle multiple levels of attacks. This simplifies the application of the model and drastically reduces the computational costs, especially in the context of transfer learning across multiple downstream tasks.

We pre-train our model on the ImageNet dataset, and transfer it to CIFAR10, CIFAR100, and some other downstream datasets.
Compared to existing works on randomized smoothing, our method achieves similar or even superior results with just a single model, as opposed to their use of several models.
Remarkably, our single model simultaneously achieved a clean accuracy of 98.0\% and a certified accuracy of 80.1\% at a radius of $\varepsilon = 0.25$ on CIFAR10, representing a 5.2\% improvement over the previous best method (see Table \ref{tab:comparison}).
Additionally, we perform extensive experiments to investigate the correlation between semantic learning and robustness during the model's transfer process.
We hope that these analyses will stimulate more exploration of robustness in transfer learning.

\begin{figure*}  
  \centering
  \includegraphics[width=1.0\textwidth]{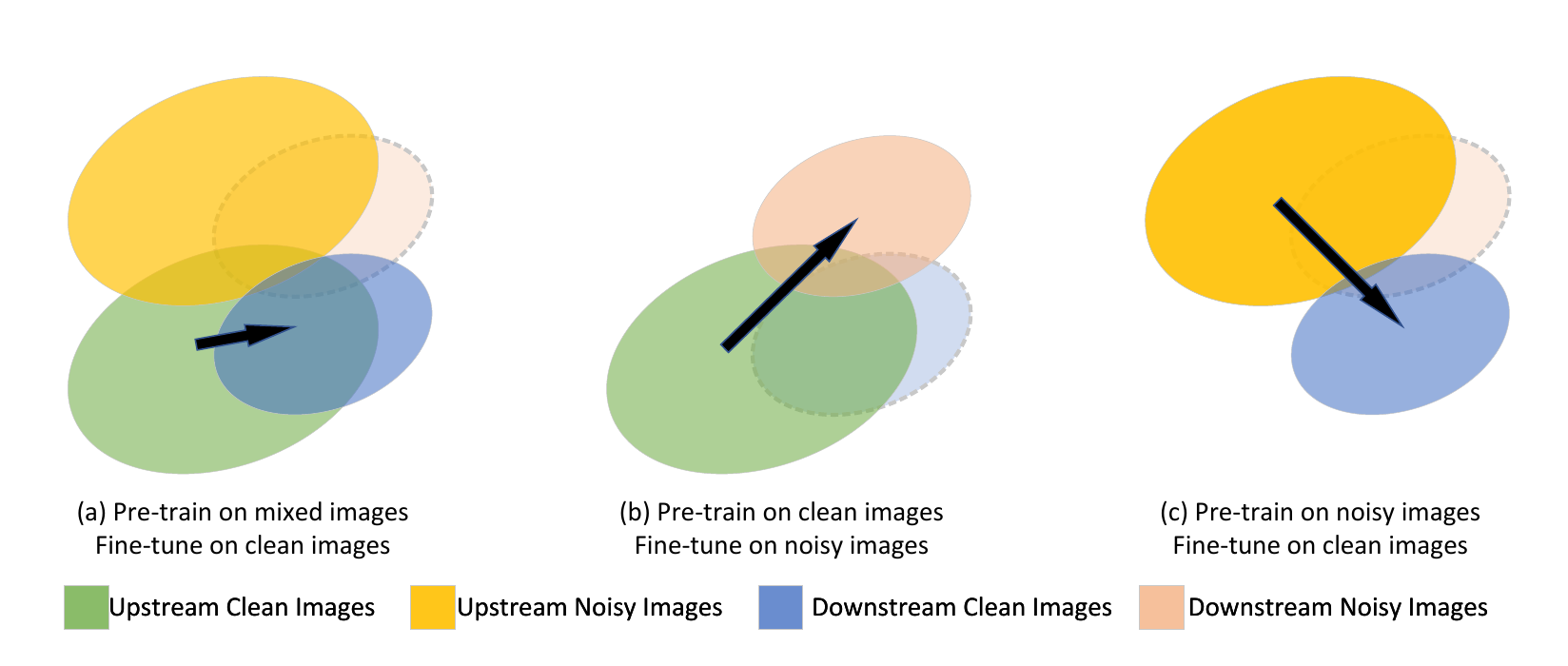}   
  \caption{Illustration of distribution gap.
  (a) The model learns a broad data distribution on a mixed dataset containing both clean and noisy data. Only minor adjustments are needed for the model to adapt to downstream clean images, hence the robustness can be retained.
  (b) When shifting from upstream clean images to downstream noisy images, the model needs to make significant adjustments to adapt to the new distribution.
  (c) The model shifts from upstream noisy images to downstream clean images. Due to the large distribution gap, the model forgets its robustness while adapting to clean images.
   }  
  \label{fig:transfer}  
\end{figure*}

\section{Related Work}
\label{sec:related_work}

Neural networks are widely observed to be fragile under imperceptible adversarial attacks, initially in the computer vision domain \cite{szegedy2013intriguing,biggio2013evasion,goodfellow2014explaining} and nowadays also in large language models  \cite{zou2023universal, maus2023adversarial, shin2020autoprompt,robey2023smoothllm}.
The topic of adversarial attacks has drawn widespread attention from researchers and practitioners.
Meanwhile, many defense approaches have been proposed. Empirically effective methods including adversarial training \cite{madry2018towards} have often been demonstrated to be defeatable with stronger attacks \cite{athalye2018obfuscated}. Therefore, defenses with provable robustness guarantees have attracted extensive study. One widely used and scalable certified robustness approach is randomized smoothing \cite{cohen2019certified, lecuyer2019certified, salman2019provably}. 

\textbf{Randomized Smoothing.} Randomized smoothing adds Gaussian noises to the input data, which theoretically makes the classifier smooth to resist any bounded attacks. A basic solution is introduced in \cite{cohen2019certified} with follow-up improved variants presented in \cite{salman2019provably,zhai2020macer,jeong2020consistency} and an ensemble approach in \cite{horvath2021boosting}. 
Recently, diffusion models have been leveraged to provide certified robustness \cite{carlini2022certified,lee2021provable}, to perform adversarial purification \cite{nie2022diffusion}, or both \cite{zhang2023diffsmooth}, as they are trained with a denoising objective and naturally induces robustness to Gaussian noises. They all require powerful diffusion models trained with extremely large datasets. Instead, we focus on exploring how to extend the successful standard pipeline of pre-training plus lightweight fine-tuning to certified robustness in general.

\textbf{Robustness and Semantic Transfer.} Our paper is also closely related to the concept of robustness transfer from a pre-trained model to a final fine-tuned model for specific tasks, which is studied in \cite{shafahi2019adversarially} for adversarial training and in \cite{salman2020denoised, wu2022denoising} for randomized smoothing. However, they focus only on robust accuracy while compromising accuracy on clean input data, thus disregarding the balance between semantic learning and robustness transfer. 
In contrast, our method can achieve both clean and certified accuracy.


\section{Method}
\label{sec:method}

We first provide a brief overview of the concepts underlying randomized smoothing, then move on to a comprehensive explanation of our innovative methodology.

\begin{table*}[]
  \centering
  \caption{The impact of normalizations on the robustness transfer. We categorize normalization into two types: Batch Normalization and Batch-independent Normalizations (including Instance Normalization, Group Normalization and Layer Normalization). In the context of mixed-noise pre-training and clean image fine-tuning, these two types of normalizations differ in the way they compute statistics. Batch Normalization uses different statistics in the pre-training and fine-tuning stages, while batch-independent normalization remains consistent.}
  \label{tab:normalization_equation}
  \begin{tabular}{ll|ll}
  \hline
  \multicolumn{2}{c|}{Batch Normalization}                      & \multicolumn{2}{c}{Batch-independent Normalization}                                 \\ \hline
  \multicolumn{1}{l|}{\multirow{4}{*}{Pre-train}} & $E_{mix} = mean(x_{clean}, x_{noise})$   & \multicolumn{1}{l|}{\multirow{4}{*}{Pre-train}} & $E_{clean} = mean(x_{clean})$, $E_{noise} = mean(x_{noise})$ \\
  \multicolumn{1}{l|}{}                              & $V_{mix} = var(x_{clean}, x_{noise})$    & \multicolumn{1}{l|}{}                              & $V_{clean} = var(x_{clean})$, $V_{noise} = var(x_{noise})$ \\
  \multicolumn{1}{l|}{}                              & ${x'}_{clean} = \frac{x_{clean} - E_{mix}}{\sqrt{V_{mix}}} \times \gamma + \beta$ & \multicolumn{1}{l|}{}                              & ${x'}_{clean} = \frac{x_{clean} - E_{clean}}{\sqrt{V_{clean}}} \times \gamma + \beta$           \\
  \multicolumn{1}{l|}{}                              & ${x'}_{noise} = \frac{x_{noise} - E_{mix}}{\sqrt{V_{mix}}} \times \gamma + \beta$ & \multicolumn{1}{l|}{}                              & ${x'}_{noise} = \frac{x_{noise} - E_{noise}}{\sqrt{V_{noise}}} \times \gamma + \beta$           \\ \hline
  \multicolumn{1}{l|}{\multirow{3}{*}{Fine-tune}}    & $E_{clean} = mean(x_{clean})$ & \multicolumn{1}{l|}{\multirow{3}{*}{Fine-tune}}    & $E_{clean} = mean(x_{clean})$           \\
  \multicolumn{1}{l|}{}                              & $V_{clean} = var(x_{clean})$ & \multicolumn{1}{l|}{}                              & $V_{clean} = var(x_{clean})$           \\
  \multicolumn{1}{l|}{}                              & ${x'}_{clean} = \frac{x_{clean} - E_{clean}}{\sqrt{V_{clean}}} \times \gamma + \beta$ & \multicolumn{1}{l|}{}                              & ${x'}_{clean} = \frac{x_{clean} - E_{clean}}{\sqrt{V_{clean}}} \times \gamma + \beta$           \\ \hline
  \end{tabular}
  \end{table*}

\subsection{Background}

Randomized smoothing, initially introduced in \cite{lecuyer2019certified, li2018second}, can convert any arbitrary base classifier $f$ into a new smoothed classifier $g$ that is certifiably robust in $\ell_2$ norm.
Specifically, the classifier $f$ maps input images $x$ to classes $\mathcal{Y}$. The smoothed classifier $g$ returns
the class that is most likely to be returned by the base classifier $f$ when $x$ is perturbed by isotropic Gaussian noise:

\begin{equation}
\begin{split}  
  g(x) = \ &\underset{c \in \mathcal{Y}}{arg\ max}\ \mathbb{P} [f(x + \delta ) = c] 
  \\ &where \ \delta \sim \mathcal{N}(0, \sigma^2I) \ .
  \label{eq:important}
\end{split}  
\end{equation}
The $\delta$ denotes Gaussian noise with standard deviation $\sigma$.

Cohen \etal proved a tight robustness guarantee in $\ell_2$ norm for the smoothed classifier \cite{cohen2019certified}.
Suppose that when the base classifier $f$ classifies $(x + \delta)$, the most likely class is $c_A$ with probability of $p_A$ and the second most likely class is $c_B$ with probability of $p_B$.
Cohen \etal applied the Neyman-Pearson lemma to prove that the smoothed classifier $g$ is robust in $\ell_2$ norm with radius $R$:
\begin{equation}
  \label{eq:radius}
  R = \frac{\sigma}{2} (\Phi^{-1}(p_A) - \Phi^{-1}(p_B)) \ ,
\end{equation}
where $\Phi^{-1}$ is the inverse of the cumulative distribution function of the standard Gaussian CDF.
The $p_A$ and $p_B$ are estimated using Monte Carlo sampling.
In order to withstand adversarial attacks of different intensities, existing works train multiple models, each with a specific $\sigma$.
To assess the overall robustness, the performance of the model under various sigma values is combined.

\subsection{Pipeline}
Our target is to pre-train a model that can easily transfer semantic features and robustness to downstream tasks, while simultaneously endeavoring to alleviate the computational expenses of fine-tuning.
Our strategy involves initially extending the data distribution on the upstream dataset, enabling the model to learn both robust and strong semantic features. We use mixed-noise training to achieve this goal. To simplify the fine-tuning process, we only train on clean images from the downstream task.
Our mixed-noise pre-training approach coupled with fine-tuning on clean images from downstream tasks has multiple benefits.
Firstly, a pre-trained model can transfer robustness to multiple downstream tasks, eliminating the need to learn robustness separately for each task. 
Secondly,  a single trained model can handle various levels of noise, negating the need for multiple models and reducing application complexity.
Thirdly, mixed-noise training allows the model to learn both semantics and robustness without having to compromise one for the other. Fourthly, using only clean images from the downstream task simplifies the fine-tuning process greatly. Finally, we found that this approach yields very good clean and certified accuracy across multiple downstream tasks.



Our pipeline consists of two steps. During the pre-training phase, we train with a mix of clean images and various different noises.
 Specifically, we construct a set $S=\{\sigma_0, \sigma_1, …, \sigma_n, …\}$ that contains a variety of noise intensities. 
During training, we randomly sample $\sigma$ from set $S$ and add it to the clean image. Remarkably, $\sigma_0=0$ represents direct training with the clean image.
Since clean images (i.e., $\sigma = 0$) occur randomly during the training process, the model is effectively able to learn the semantic representation.
Incorporating noisy images into the training set forces the model's decision boundary to include both clean and noisy representations of the same category. This enables the model to learn robustness.
During fine-tuning on downstream datasets, we only train on clean images.
Even so, we still achieved strong clean and certified accuracy.
Given that only clean images are used during fine-tuning, the model's robustness must be derived from the pre-training phase.
In some more severe scenarios, we found that even just one epoch of fine-tuning can yield excellent clean accuracy and certified robustness on the downstream dataset (see Table \ref{tab:sum}).

\subsection{Distribution Gap}
We found that to ensure robustness while transferring semantics, it is crucial to minimize the distribution gap between the pre-training and fine-tuning datasets (see Figure \ref{fig:transfer}).
If the distribution gap between the pre-training and fine-tuning datasets is small, the model only needs to make minor adjustments on the distribution to adapt to the new dataset. Thus, both semantic representation and robustness can be transferred simultaneously.
However, if the distribution gap between the pre-training and fine-tuning datasets is large, the model will need to make a substantial adjustment to adapt to the new dataset. This forces the model to make trade-offs between semantic representation and robustness.

The set of noises used in the pre-training and fine-tuning stages has a significant impact on the distribution difference.
If the set of noises used in fine-tuning stages is covered in the pre-training stage, the distribution gap is small. On the other hand, if there is no overlap between the sets of noises used in these two stages, the distribution gap becomes large and the model suffers from a compromise between robustness and semantic transfer.
Mixing various levels of noise during the pre-training phase allows the model to learn a broad distribution, reducing the disparity with the distribution of the downstream datasets.
Therefore, even if only clean images from downstream are used, semantic features and robustness can still be well transferred.

\subsection{Statistics Adaptation}

The ability to simultaneously transfer the upstream semantics and robustness to downstream tasks depends on a variety of factors.
We found that normalization is a key factor.
The majority of the current randomized smoothing research predominantly uses ResNet for experiments. 
However, we discovered that Batch Normalization in ResNet significantly undermines the transferability of model robustness.
We provide a detailed analysis of common normalizations in Table \ref{tab:normalization_equation}.
Batch Normalization calculates statistics (i.e., mean and variance) across multiple instances, whereas other batch-independent normalizations only calculate within a single instance.
When the upstream and downstream distributions are inconsistent, Batch Normalization uses completely different statistics in the pre-training and fine-tuning stages, leading to a significant distribution shift. In contrast, other batch-independent normalizations use consistent statistics in the upstream and downstream, ensuring a smooth transfer.
In our experiments, we found that Batch Normalization has to compromise between clean and certified accuracy. 
In contrast, other normalizations can transfer both semantics and robustness simultaneously (see Table \ref{tab:norm}).




\begin{table*}[t]
  \centering
  \caption{Comparison with existing work on the CIFAR10 dataset. Each method uses three noise levels $\sigma \in \{0.25, 0.5, 1.0\}$ to produce three certified accuracy curves (see Figure \ref{fig:vis}).
  The maximum values of these curves at given radius $\varepsilon \in \{0.25, 0.5, 0.75, 1.0 \}$ are reported in this table. 
  The numbers enclosed in brackets indicate the certified accuracy of the respective curve at $\varepsilon=0$. Beyond certified accuracy, we also report the clean accuracy achieved by our method. While most of the existing methods compromise clean accuracy for certified accuracy,  our method can achieve strong performance in both clean and certified accuracy.}
  \label{tab:comparison}
  \begin{tabular}{cc|c|cccc}
    \hline
    \multicolumn{2}{c|}{\multirow{2}{*}{Method}} & \multirow{2}{*}{\begin{tabular}[c]{@{}c@{}}Clean Acc.\end{tabular}} & \multicolumn{4}{c}{Certified Acc. at $\varepsilon$ (\%)} \\ \cline{4-7} 
    \multicolumn{2}{c|}{}       &                                                                       & 0.25           & 0.5           & 0.75           & 1.0          \\ \hline
    \multicolumn{2}{l|}{PixelDP  \citeyear{lecuyer2019certified} \cite{lecuyer2019certified}} & -  &  $ \prescript{(71.0)}{}{22.0}$   &   $ \prescript{(44.0)}{}{2.0}$     &    -    &   -      \\
    \multicolumn{2}{l|}{Randomized Smoothing \citeyear{cohen2019certified} \cite{cohen2019certified}}  &  -  &  $ \prescript{(75.0)}{}{61.0}$   &   $ \prescript{(75.0)}{}{43.0}$     &    $ \prescript{(65.0)}{}{32.0}$    &   $ \prescript{(66.0)}{}{22.0}$      \\
    \multicolumn{2}{l|}{SmoothAdv \citeyear{salman2019provably} \cite{salman2019provably}} &  -  &  $ \prescript{(84.3)}{}{74.9}$   &   $ \prescript{(80.1)}{}{63.4}$     &    $ \prescript{(80.1)}{}{51.9}$    &   $ \prescript{(62.2)}{}{39.6}$      \\
    \multicolumn{2}{l|}{Consistency \citeyear{jeong2020consistency} \cite{jeong2020consistency}} &  -  &  $ \prescript{(77.8)}{}{68.8}$   &   $ \prescript{(75.8)}{}{58.1}$     &    $ \prescript{(72.9)}{}{48.5}$    &   $ \prescript{(52.3)}{}{37.8}$      \\
    \multicolumn{2}{l|}{MACER \citeyear{zhai2020macer} \cite{zhai2020macer} }  & -   &  $ \prescript{(81.0)}{}{71.0}$   &   $ \prescript{(81.0)}{}{59.0}$     &    $ \prescript{(66.0)}{}{46.0}$    &   $ \prescript{(66.0)}{}{38.0}$      \\
    \multicolumn{2}{l|}{Denoised \citeyear{salman2020denoised} \cite{salman2020denoised} }  & -   &  $ \prescript{(72.0)}{}{56.0}$   &   $ \prescript{(62.0)}{}{41.0}$     &    $ \prescript{(62.0)}{}{28.0}$    &   $ \prescript{(44.0)}{}{19.0}$      \\
    \multicolumn{2}{l|}{Boosting \citeyear{horvath2021boosting} \cite{horvath2021boosting} }  & -   &  $ \prescript{(83.4)}{}{70.6}$   &   $ \prescript{(76.8)}{}{60.4}$     &    $ \prescript{(71.6)}{}{52.4}$    &   $ \prescript{(52.4)}{}{38.8}$      \\
    \multicolumn{2}{l|}{DRT \citeyear{yang2021certified} \cite{yang2021certified} }  & -   &  $ \prescript{(81.5)}{}{70.4}$   &   $ \prescript{(72.6)}{}{60.2}$     &    $ \prescript{(71.9)}{}{50.5}$    &   $ \prescript{(56.1)}{}{39.8}$      \\
    \multicolumn{2}{l|}{SmoothMix \citeyear{jeong2021smoothmix} \cite{jeong2021smoothmix} }  & -   &  $ \prescript{(77.1)}{}{67.9}$   &   $ \prescript{(77.1)}{}{57.9}$     &    $ \prescript{(74.2)}{}{47.7}$    &   $ \prescript{(61.8)}{}{37.2}$      \\
    \multicolumn{2}{l|}{Lee \citeyear{lee2021provable} \cite{lee2021provable} }  & -   &  60.0   &   42.0     &    28.0    &   19.0      \\
    \multicolumn{2}{l|}{ACES \citeyear{horvath2022robust} \cite{horvath2022robust} }  & -   &  $ \prescript{(79.0)}{}{69.0}$   &   $ \prescript{(74.2)}{}{57.2}$     &    $ \prescript{(74.2)}{}{47.0}$    &   $ \prescript{(58.6)}{}{37.8}$      \\ \hline
    \multicolumn{2}{l|}{\textbf{Ours}}  & \textbf{98.0}  &  $ \prescript{(\textbf{90.2})}{}{\textbf{80.1}}$   &   $ \prescript{(90.2)}{}{62.5}$     &    $ \prescript{(90.2)}{}{45.1}$    &   $ \prescript{(77.3)}{}{31.6}$      \\ \hline
 \end{tabular}
\end{table*}

\section{Experiments}
\label{sec:experiments}

In this section, we first introduce the experimental setup, followed by a comparison with existing work.
It is noteworthy that with only one model, our method achieves results that are on par with or even better than the best outcomes from existing methods.
Lastly, to delve into the properties of semantic learning and robustness during the transfer process, we conduct extensive ablation studies.

\begin{figure*}
  \centering
  \begin{subfigure}{0.49\linewidth}
    \includegraphics[width=1.0\textwidth]{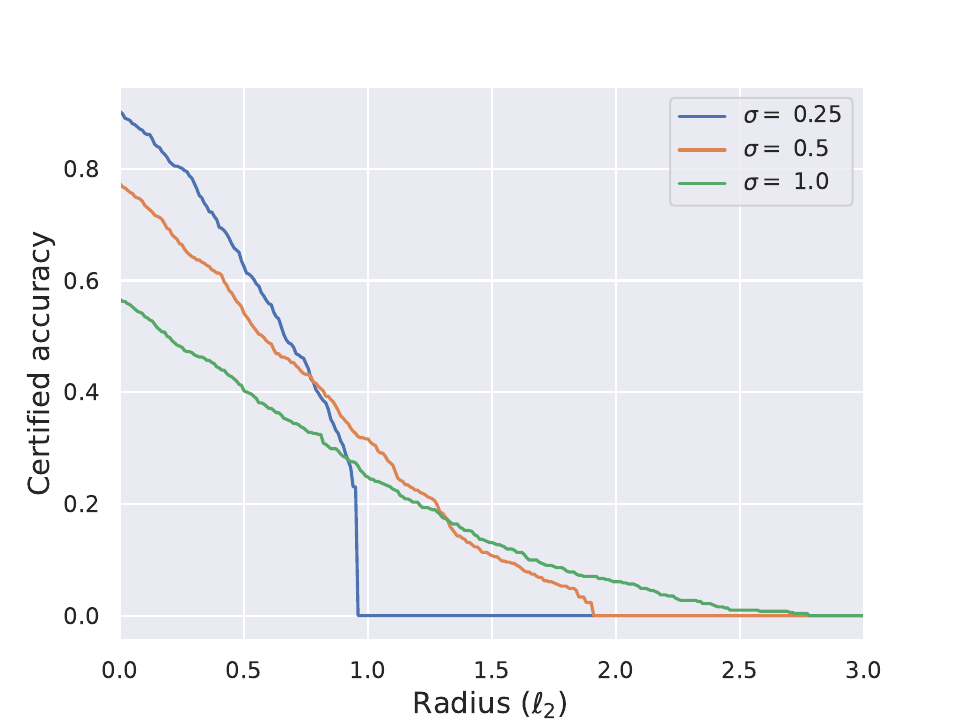}   
    \caption{CIFAR10.}
    \label{fig:cifar10 exp}
  \end{subfigure}
  \hfill
  \begin{subfigure}{0.49\linewidth}
    \includegraphics[width=1.0\textwidth]{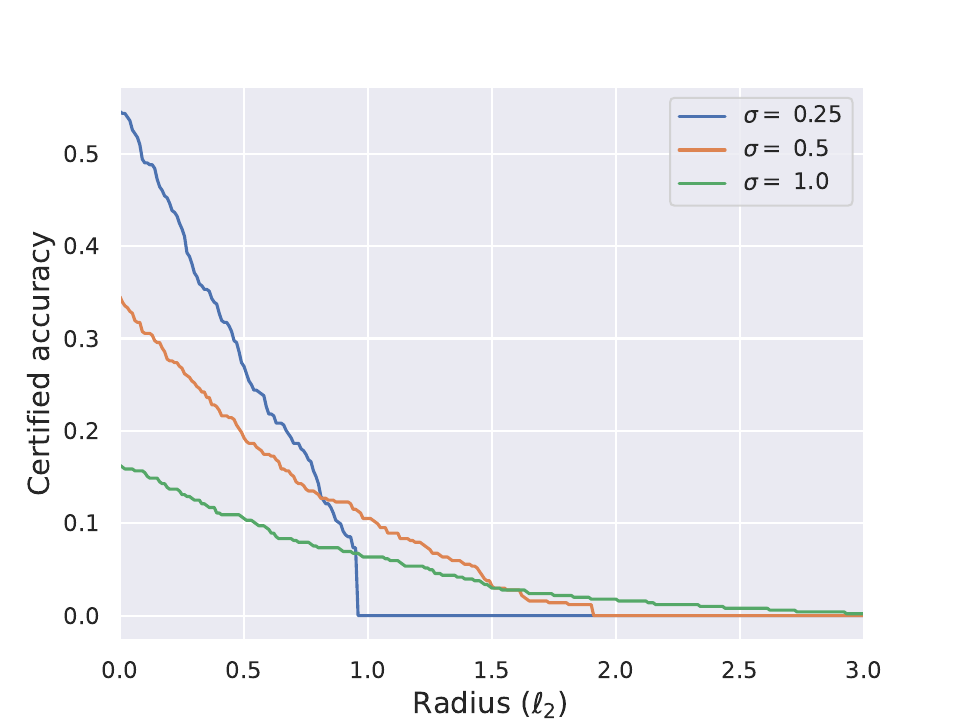}   
    \caption{CIFAR100.}
    \label{fig:cifar100 exp}
  \end{subfigure}
  \caption{Certified accuracy on CIFAR10 (left) and CIFAR100 (right). We present results with three noise levels $\sigma \in \{0.25, 0.5, 1.0\}$ for each dataset. We require only one model pre-trained on ImageNet with mixed noise. This model can be reused for multiple downstream tasks, including CIFAR10, CIFAR100, among others. Moreover, this single model can handle various levels of noise. The three lines in the figure are the results of the same model under three different levels of noise. Notably, the results on the right graph were obtained by merely fine-tuning on clean images from CIFAR100 for a single epoch. Pre-training with mixed noise enables the model to be easily transferred to downstream tasks.}
  \label{fig:vis}
\end{figure*}

\subsection{Experimental Setup}

\textbf{Datasets.}
Our primary focus is on how the model can effectively transfer the semantic features and robustness learned on the upstream datasets to the downstream tasks.
To ensure a fair comparison with existing work, we follow the setup outlined in \cite{salman2019provably}, mainly conducting our experiments on ImageNet \cite{deng2009imagenet}, CIFAR10, and CIFAR100.
Considering the different image sizes of ImageNet  and CIFAR10, CIFAR100, we follow \cite{salman2019provably} and adjust the size of ImageNet to 32x32 for pre-training.
We present a comparison with existing work on CIFAR10.
Moreover, we have conducted extensive ablation studies on CIFAR10 to examine the relationship between semantic learning and robustness in the transfer process.
Results on additional downstream datasets are provided in the supplementary materials.

\noindent \textbf{Models.}
In line with most existing work, we employ ResNet for our experiments.
ResNet heavily employs Batch Normalization for normalization.
We discover that Batch Normalization greatly damages the transferability of model robustness.
Our experiments include an extensive examination of Batch Normalization, Group Normalization, Layer Normalization, and Instance Normalization.
Ultimately, we choose to replace Batch Normalization with Layer Normalization in ResNet.
Since the capacity of the model is essential for robustness, we perform our experiments using ResNet152.

\noindent \textbf{Metrics.}
Most of the prior work utilize certified accuracy as a metric to evaluate the robustness of the model.
We follow the methodology outlined in \cite{cohen2019certified} to calculate certified accuracy.
In particular, we use three different levels of noise $(\sigma \in \{0.25, 0.5, 1.0\})$ for testing on CIFAR10 and CIFAR100.
For each image in the test set, we utilize N=100,000 Monte Carlo samples to estimate the values of $p_A$ and $p_B$ in Equation \ref{eq:radius}, and compute the corresponding radius $R$.
we use the certify probability to 0.001, so there is at most a 0.1\% chance that $R$ is not truly robust.
We present the certified accuracy using tables and figures. When using figures (see Figure \ref{fig:vis}), we draw three curves, each corresponding to $ \sigma \in \{0.25, 0.5, 1.0 \} $. When employing tables, we select certain specific radius $ \varepsilon \in \{ 0.25, 0.5, 0.75, 1.0 \} $. Additionally, we also report the accuracy on clean images in the test set.

\noindent \textbf{Other Details.}
When transferring the model from upstream to downstream datasets, we discard the last FC layer and adapt a randomly initialized FC to fit the downstream tasks.
Fine-tuning can be done in two ways. One approach is to train the full network.
The other is to keep the features fixed and train only the final FC layer (see Table \ref{tab:sum}).
We use eight V100 GPUs to train our models. Unless otherwise specified, we consistently implement the following training strategy.
The batch size is configured to 128, and the number of epochs is set to 100. Three learning rates (0.1, 0.01, 0.001) are used for each model, and the best outcome is reported.
We adjust the learning rate by using a linear warmup for the first 10 epochs, followed by a cosine learning rate after the 10th epoch. 
This technique ensures the model's stable convergence. We utilize Pytorch's Distributed Data Parallel (DDP) and automatic mixed precision (AMP) to accelerate training speed.

\subsection{Comparison with Existing Methods}
Table \ref{tab:comparison} provides a comparison between our work and existing methods. These prior works train multiple models using different noises and report the best results across four radii. In contrast, we train a single model to deal with various levels of noise.
Some studies also pre-train the model on larger datasets and then fine-tune it on CIFAR10. For instance, SmoothAdv uses two datasets (ImageNet and Tiny500K) for pre-training \cite{salman2019provably}. For simplicity, we only pre-train on ImageNet. Other studies utilize even larger datasets \cite{carlini2022certified}, such as ImageNet22K. To ensure fairness, we have not included these studies in our table.
Some works employ an ensemble of many models to obtain impressive results \cite{jeong2020consistency}. Yet, we did not resort to any ensemble methods. Nevertheless, our method still outperforms the others on various metrics. At $\varepsilon=0.25$, our certified accuracy stands at 80.1\%, marking a 5.2\% increase from the previously best-reported result.
Particularly noteworthy is that most existing works often sacrifice clean accuracy in exchange for certified accuracy.
However, our method achieves good results in both clean and certified accuracy. 
The full certified accuracy of our method is depicted in Figure \ref{fig:cifar10 exp}.

\begin{table*}[t]
  \centering
  \caption{Effects of pre-training. The top four rows represent the results of training from scratch on CIFAR10.
  The last row represents pre-training on ImageNet with mixed noise, followed by fine-tuning on clean images from CIFAR10 for only one epoch.
  Models trained from scratch inevitably have to compromise between clean and certified accuracy.
  Pre-training with mixed noise allows the model to learn a broad data distribution.
  This capacity enables the model to not only transfer semantic representations to downstream tasks for impressive clean accuracy but also transfer robustness to downstream tasks for excellent certified accuracy.}
  \label{tab:pretraining}
  \begin{tabular}{cc|c|cccc}
  \hline
  \multicolumn{2}{c|}{\multirow{2}{*}{Method}}                     & \multirow{2}{*}{Clean Acc.} & \multicolumn{4}{c}{Certified Acc. at $\varepsilon$ (\%)}           \\ \cline{4-7} 
  \multicolumn{2}{c|}{}                                            &                             & \multicolumn{1}{c}{0.25} & 0.5 & 0.75 & 1.0 \\ \hline
  \multicolumn{1}{c|}{\multirow{4}{*}{Train from scratch}} & Clean & 87.5  &  $ \prescript{(10.5)}{}{10.5}$   &   $ \prescript{(10.5)}{}{10.1}$     &    $ \prescript{(10.5)}{}{9.7}$    &   $ \prescript{(10.5)}{}{5.5}$      \\
  \multicolumn{1}{c|}{}                                    & $\sigma = 0.25$  & 71.8  &  $ \prescript{(74.2)}{}{59.3}$   &   $ \prescript{(74.2)}{}{41.7}$     &    $ \prescript{(74.2)}{}{26.0}$    &   $ \prescript{(46.0)}{}{6.7}$      \\
  \multicolumn{1}{c|}{}                                    & $\sigma = 0.5$   & 54.9  &  $ \prescript{(61.9)}{}{49.0}$   &   $ \prescript{(61.9)}{}{38.7}$     &    $ \prescript{(61.9)}{}{\textbf{29.6}}$    &   $ \prescript{(61.9)}{}{\textbf{20.8}}$      \\
  \multicolumn{1}{c|}{}                                    & $\sigma = 1.0$   & 41.1  &  $ \prescript{(45.0)}{}{39.1}$   &   $ \prescript{(45.0)}{}{30.0}$     &    $ \prescript{(45.0)}{}{24.8}$    &   $ \prescript{(40.9)}{}{19.8}$      \\ \hline
  \multicolumn{2}{c|}{Mixed noise pre-train + clean fine-tune $\star$}                                 & \textbf{96.8}  &  $ \prescript{(\textbf{81.3})}{}{\textbf{66.7}}$   &   $ \prescript{(\textbf{81.3})}{}{\textbf{45.0}}$     &    $ \prescript{(81.3)}{}{27.2}$    &   $ \prescript{(60.9)}{}{16.9}$      \\ \hline
  \end{tabular}
\end{table*}

\begin{table*}[t]
  \centering
  \caption{Effects of mixed noise.
  All these experiments involve pre-training on ImageNet, followed by finetuning on CIFAR10.
  The difference lies in the noise used during the training process.
  Models that are solely trained on clean images might demonstrate high clean accuracy, but their certified accuracy nearly approximates random guessing.
  Models trained under specific noise exhibit good performance on certain certified accuracy metrics, but at the cost of clean accuracy.
  In contrast, mixed noise attains the best, or near-best, performance across all metrics.
  }
  \label{tab:single_mixed}
  \begin{tabular}{cc|c|cccc}
    \hline
    \multicolumn{2}{c|}{\multirow{2}{*}{Pre-training and fine-tuning}} & \multirow{2}{*}{\begin{tabular}[c]{@{}c@{}}Clean Acc.\end{tabular}} & \multicolumn{4}{c}{Certified Acc. at $\varepsilon$ (\%)} \\ \cline{4-7} 
    \multicolumn{2}{c|}{}       &                                                                       & 0.25           & 0.5           & 0.75           & 1.0          \\ \hline
    \multicolumn{2}{c|}{Clean} & 97.8  &  $ \prescript{(10.1)}{}{10.1}$   &   $ \prescript{(10.1)}{}{10.1}$     &    $ \prescript{(10.1)}{}{10.1}$    &   $ \prescript{(10.1)}{}{10.1}$      \\
    \multicolumn{2}{c|}{$\sigma =$ 0.25}  &  82.7  &  $ \prescript{(\textbf{89.7})}{}{\textbf{78.6}}$   &   $ \prescript{(89.7)}{}{57.1}$     &    $ \prescript{(\textbf{89.7})}{}{\textbf{38.5}}$    &   $ \prescript{(46.6)}{}{11.7}$      \\
    \multicolumn{2}{c|}{$\sigma =$ 0.5} &  64.1  &  $ \prescript{(74.2)}{}{62.3}$   &   $ \prescript{(74.2)}{}{48.8}$     &    $ \prescript{(74.2)}{}{36.1}$    &   $ \prescript{(\textbf{74.2})}{}{\textbf{25.2}}$      \\
    \multicolumn{2}{c|}{$\sigma =$ 1.0} &  38.8  &  $ \prescript{(52.2)}{}{42.1}$   &   $ \prescript{(52.2)}{}{34.1}$     &    $ \prescript{(52.2)}{}{27.4}$    &   $ \prescript{(52.2)}{}{22.0
    }$      \\
    \multicolumn{2}{c|}{Mixed noise}  & \textbf{97.9}  &  $ \prescript{(88.7)}{}{77.6}$   &   $ \prescript{(88.7)}{}{\textbf{57.5}}$     &    $ \prescript{(88.7)}{}{36.7}$    &   $ \prescript{(72.6)}{}{23.8}$      \\ \hline
    \end{tabular}
\end{table*}

\subsection{Ablation Study}

\textbf{Effects of pre-training.}
Table \ref{tab:pretraining} offers a comparative view of experiments involving training from scratch against those deploying pre-training paired with fine-tuning.
Models trained from scratch inevitably have to compromise between clean and certified accuracy. 
As $\sigma$ increases, the clean accuracy experiences a more significant drop.
This observation is in line with previous work \cite{cohen2019certified}.
Models pre-trained with mixed noise, even when fine-tuned on clean images for just one epoch, obtain impressive clean and certified accuracy. 
With a clean accuracy of 96.8\%, it surpasses the best result from training from scratch by 9.3\%.
This indicates that the semantic features learned in the pre-training phase have been successfully applied to the downstream task.
Moreover, the certified accuracy exceeds that from training from scratch at $\varepsilon$ of 0.25 and 0.5.
Since the model has not encountered noise during the fine-tuning phase, robustness can only stem from the pre-training phase.
This indicates that robustness has also been successfully transferred to the downstream task.
It is worth noting that the fine-tuning here is very lightweight. When there are multiple downstream tasks, this can significantly reduce computational costs.

\noindent
\textbf{Effects of mixed noise.}
Table \ref{tab:single_mixed} presents comparison experiments between utilizing single or mixed noise.
All of these experiments involve pre-training followed by fine-tuning, so their training costs are consistent.
The first row in the table trains only on clean images. 
Despite achieving strong clean accuracy, the certified accuracy turns out to be extremely low.
The experiments in the second to fourth rows of the table represent the standard settings used in previous work \cite{salman2019provably}.
They usually train multiple models, with each model adopting the same noise during pre-training and fine-tuning.
Each model is used to handle specific noise.
Although each model performs well under specific noise, the need for multiple models presents complications in practical use.
Compared to these, we only train a single model that can handle various noises, and it approaches or achieves the best performance across all metrics.
The ability of our single pre-trained model to accommodate multiple downstream tasks and handle various levels of noise greatly simplifies practical deployment.

\begin{table*}[]
  \centering
  \caption{Effects of normalization layer. We conduct a comparative analysis of four prevalent normalization layers to examine their effects on the transfer learning process. Our primary focus is the influence of normalization layers on transferability. We employ mixed noise during the pre-training phase on ImageNet, and then fine-tune on clean images from CIFAR10. This allows us to examine whether the model can maintain its robustness while transferring the semantic features.
  }
  \label{tab:norm}
  \begin{tabular}{cc|c|cccc}
    \hline
    \multicolumn{2}{l|}{\multirow{2}{*}{Normalization layer}} & \multirow{2}{*}{\begin{tabular}[c]{@{}c@{}}Clean Acc.\end{tabular}} & \multicolumn{4}{c}{Certified Acc. at $\varepsilon$ (\%)} \\ \cline{4-7} 
    \multicolumn{2}{c|}{}       &                                                                       & 0.25           & 0.5           & 0.75           & 1.0          \\ \hline
    \multicolumn{2}{l|}{Batch Normalization} & 98.3  &  $ \prescript{(18.1)}{}{12.3}$   &   $ \prescript{(12.7)}{}{8.7}$     &    $ \prescript{(8.3)}{}{8.3}$    &   $ \prescript{(8.3)}{}{8.3}$      \\
    \multicolumn{2}{l|}{Instance Normalization}  &  97.6  &  $ \prescript{(83.9)}{}{66.3}$   &   $ \prescript{(83.9)}{}{41.5}$     &    $ \prescript{(83.9)}{}{21.8}$    &   $ \prescript{(40.9)}{}{4.8}$      \\
    \multicolumn{2}{l|}{Group Normalization} &  98.3  &  $ \prescript{(87.7)}{}{71.4}$   &   $ \prescript{(87.7)}{}{48.2}$     &    $ \prescript{(87.7)}{}{28.8}$    &   $ \prescript{(45.6)}{}{8.3}$      \\
    \multicolumn{2}{l|}{Layer Normalization} &  98.0  &  $ \prescript{(85.5)}{}{71.4}$   &   $ \prescript{(85.5)}{}{49.0}$     &    $ \prescript{(85.5)}{}{25.4}$    &   $ \prescript{(39.1)}{}{17.7}$      \\
    \hline
    \end{tabular}
\end{table*}

\begin{table*}[]
  \centering
  \caption{Experiments on CIFAR100. The top four rows represent the results of training from scratch on CIFAR100. Training on clean images can yield relatively good clean accuracy, but the certified accuracy is very poor. Training with noise at different $\sigma$ can achieve decent certified accuracy, but it comes at the cost of sacrificing clean accuracy. The last row represents pre-training on ImageNet with mixed noise, followed by fine-tuning on clean images from CIFAR100 for one epoch. Through this simple fine-tuning,  the semantic features and robustness from the pre-training phase can be simultaneously transferred to the downstream task.}
  \label{tab:cifar100}
  \begin{tabular}{cc|c|cccc}
  \hline
  \multicolumn{2}{c|}{\multirow{2}{*}{Method}}                     & \multirow{2}{*}{Clean Acc.} & \multicolumn{4}{c}{Certified Acc. at $\varepsilon$ (\%)}           \\ \cline{4-7} 
  \multicolumn{2}{c|}{}                                            &                             & \multicolumn{1}{c}{0.25} & 0.5 & 0.75 & 1.0 \\ \hline
  \multicolumn{1}{c|}{\multirow{4}{*}{Train from scratch}} & Clean & 63.7    & $ \prescript{(3.2)}{}{1.4}$      & $ \prescript{(3.2)}{}{0.8}$          & $ \prescript{(0.6)}{}{0.6}$  &    $ \prescript{(0.6)}{}{0.6}$         \\
  \multicolumn{1}{c|}{}                                    & $\sigma = 0.25$  & 29.5  &  $ \prescript{(47.5)}{}{33.6}$   &   $ \prescript{(47.5)}{}{22.8}$     &    $ \prescript{(47.5)}{}{12.4}$    &   $ \prescript{(20.9)}{}{3.2}$      \\
  \multicolumn{1}{c|}{}                                    & $\sigma = 0.5$   & 22.3  &  $ \prescript{(36.0)}{}{27.5}$   &   $ \prescript{(36.0)}{}{22.1}$     &    $ \prescript{(36.0)}{}{15.8}$    &   $ \prescript{(36.0)}{}{\textbf{11.5}}$      \\
  \multicolumn{1}{c|}{}                                    & $\sigma = 1.0$   & 14.29  &  $ \prescript{(18.1)}{}{15.2}$   &   $ \prescript{(18.7)}{}{13.1}$     &    $ \prescript{(18.7)}{}{10.7}$    &   $ \prescript{(18.7)}{}{8.8}$      \\ \hline
  \multicolumn{2}{c|}{Mixed noise pre-train + clean fine-tune $\star$}                                 & \textbf{79.4}  &  $ \prescript{(\textbf{54.6})}{}{\textbf{41.9}}$   &   $ \prescript{(\textbf{54.6})}{}{\textbf{27.0}}$     &    $ \prescript{(\textbf{54.6})}{}{\textbf{17.5}}$    &   $ \prescript{(34.5)}{}{10.5}$      \\ \hline
  \end{tabular}
\end{table*}

\noindent
\textbf{Effects of normalization layer.}
Popular normalization procedures include Batch Normalization, Instance Normalization, Group Normalization, and Layer Normalization.
To systematically analyze the impact of normalization on transfer, we replace the default Batch Normalization in ResNet with each of the other three normalization layers, while keeping the rest of the network unchanged.
For Group Normalization, we set the number of groups to 32.
To minimize their differences, we set the learnable parameters of these four normalization layers to be equal to the number of feature channels. Therefore, the only difference between these four normalization layers is the method for calculating the statistics (i.e., mean and variance).
Among these four normalization layers, only Batch Normalization is related to the batch dimension, calculating statistics across multiple instances. 
The remaining three normalizations are batch-independent, computing statistics within single instances.

As seen from Table \ref{tab:norm}, all four types of normalization have achieved excellent clean accuracy in downstream tasks. However, Batch Normalization performs very poorly in terms of certified accuracy.
The other three batch-independent normalizations achieve good certified accuracy.
We think this is due to the large discrepancy in the statistics of Batch Normalization between the upstream and downstream data.
Let us discuss in detail the difference when normalizing a clean image feature upstream and downstream. During the upstream mixed-noise training, the statistics used to normalize this feature are derived from multiple instances (including clean and different noises). However, during downstream fine-tuning, the statistics are only obtained from clean images.
The feature of a clean image is normalized using completely different statistics in the upstream and downstream training.
The discrepancy in the statistics results in a substantial difference in the distribution of the feature space.
In order to adapt to the downstream task, the model must make significant adjustments, thus forgetting the robustness learned in the upstream task.
In contrast, the other three batch-independent normalizations use statistics derived from a single instance, which is consistent between upstream and downstream data.

Existing randomized smoothing techniques predominantly utilize ResNet. 
The Batch Normalization, which is used by default in ResNet, has a crucial impact on the transfer of robustness.
This is an important but easily overlooked issue.
Shafahi \etal mentioned that in order to transfer robustness, the features need to be fixed and only the last FC layer should be trained during fine-tuning \cite{shafahi2019adversarially}.
However, our findings indicate that robustness can be transferred just by replacing Batch Normalization with other batch-independent normalizations. 
In fact, when we used Layer normalization in our experiments, fine-tuning the full network achieved better results than keeping the features fixed (see Table \ref{tab:sum}).
We recommend that when performing robustness transfer, Batch Normalization should be abandoned in favor of other batch-independent normalizations.
In this paper, unless otherwise specified, we use Layer normalization.

\subsection{Experiments on Other Downstream Datasets.}
Our methodology only requires the pre-training of one model which can be used across multiple downstream tasks.
We demonstrate this in Table \ref{tab:cifar100} with experiments on CIFAR100.
Training from scratch directly on CIFAR100 inevitably results in a compromise between clean and certified accuracy.
The certified accuracy is notably poor when trained solely on clean images. Conversely, when trained on noisy images at various $\sigma$, clean accuracy is sacrificed.
In contrast, the model pre-trained on ImageNet with mixed noise only needs to be fine-tuned on the clean images of CIFAR100 for one epoch to simultaneously transfer semantic features and robustness to the downstream task. This approach does not require a trade-off between clean and certified accuracy.
It not only obtains a very high clean accuracy but also attains the best performance on certified accuracy with $\varepsilon$ at 0.25, 0.5 and 0.75.
The comprehensive certified accuracy of this experiment is presented in Figure \ref{fig:cifar100 exp}.
For experimental results on other downstream datasets, please refer to the supplementary materials.

\section{Conclusion}
\label{sec:conclusion}
We propose a method to pretrain certifiably robust models that can easily transfer to downstream tasks. A key challenge is to resolve the trade-off between semantic learning and robustness. We tackle this problem by expanding the data distribution during the pretraining phase. 
Through mixed noise training, our model can simultaneously transfer semantics and robustness to downstream tasks. 
Even just fine-tuning on downstream clean images can achieve high certified accuracy.
Moreover, our single model can handle multiple types of noise, offering great practical value.

{
    \small
    \bibliographystyle{ieeenat_fullname}
    \bibliography{main}
}

\clearpage
\setcounter{page}{1}

\appendix

\section*{Supplementary Materials}

\begin{table*}[t]
  \centering
  \caption{Experiments on Pets. The first four rows represent the results of training with only clean images or specific noise. The last row represents pre-training on ImageNet with mixed noise, followed by fine-tuning on clean images from Pets. Through this simple fine-tuning, it achieves the best clean and certified accuracy.
  }
  \label{tab:pets}
  \begin{tabular}{cc|c|cccc}
    \hline
    \multicolumn{2}{l|}{\multirow{2}{*}{Method}} & \multirow{2}{*}{\begin{tabular}[c]{@{}c@{}}Clean Acc.\end{tabular}} & \multicolumn{4}{c}{Certified Acc. at $\varepsilon$ (\%)} \\ \cline{4-7} 
    \multicolumn{2}{c|}{}       &                                                                       & 0.25           & 0.5           & 0.75           & 1.0          \\ \hline
    \multicolumn{2}{l|}{Clean} & 91.3    & $ \prescript{(6.0)}{}{3.5}$      & $ \prescript{(3.5)}{}{3.3}$          & $ \prescript{(3.5)}{}{3.0}$  &    $ \prescript{(3.5)}{}{2.2}$         \\
    \multicolumn{2}{l|}{$\sigma = 0.25$}  &  50.8  &  $ \prescript{(71.2)}{}{60.3}$   &   $ \prescript{(71.2)}{}{49.7}$     &    $ \prescript{(71.2)}{}{39.7}$    &   $ \prescript{(2.4)}{}{2.4}$      \\
    \multicolumn{2}{l|}{$\sigma = 0.5$} &  36.2  &  $ \prescript{(56.0)}{}{50.3}$   &   $ \prescript{(56.0)}{}{45.7}$     &    $ \prescript{(56.0)}{}{39.9}$    &   $ \prescript{(56.0)}{}{35.1}$      \\
    \multicolumn{2}{l|}{$\sigma = 1.0$} &  24.6  &  $ \prescript{(40.5)}{}{38.0}$   &   $ \prescript{(40.5)}{}{35.3}$     &    $ \prescript{(40.5)}{}{31.5}$    &   $ \prescript{(40.5)}{}{26.6}$      \\ \hline
    \multicolumn{2}{l|}{\textbf{Mixed Noise}} & \textbf{91.8} &  $ \prescript{(\textbf{88.6})}{}{\textbf{84.2}}$   &   $ \prescript{(\textbf{88.6})}{}{\textbf{75.5}}$     &    $ \prescript{(\textbf{85.1})}{}{\textbf{65.2}}$    &   $ \prescript{(\textbf{85.1})}{}{\textbf{57.1}}$      \\
    \hline
    \end{tabular}
\end{table*}

\begin{table*}[h]
  \centering
  \caption{Experiments on Food101. Mixed noise pre-training enables the model to achieve the best clean and certified accuracy.
  }
  \label{tab:Food101}
  \begin{tabular}{cc|c|cccc}
    \hline
    \multicolumn{2}{l|}{\multirow{2}{*}{Method}} & \multirow{2}{*}{\begin{tabular}[c]{@{}c@{}}Clean Acc.\end{tabular}} & \multicolumn{4}{c}{Certified Acc. at $\varepsilon$ (\%)} \\ \cline{4-7} 
    \multicolumn{2}{c|}{}       &                                                                       & 0.25           & 0.5           & 0.75           & 1.0          \\ \hline
    \multicolumn{2}{l|}{Clean} & 78.0    & $ \prescript{(6.4)}{}{4.0}$      & $ \prescript{(6.4)}{}{2.5}$          & $ \prescript{(6.4)}{}{0.9}$  &    $ \prescript{(0.9)}{}{0.8}$         \\
    \multicolumn{2}{l|}{$\sigma = 0.25$}  &  56.0  &  $ \prescript{(71.4)}{}{61.7}$   &   $ \prescript{(71.4)}{}{51.7}$     &    $ \prescript{(71.4)}{}{38.4}$    &   $ \prescript{(44.5)}{}{20.0}$      \\
    \multicolumn{2}{l|}{$\sigma = 0.5$} &  42.9  &  $ \prescript{(62.8)}{}{56.3}$   &   $ \prescript{(62.8)}{}{50.0}$     &    $ \prescript{(62.8)}{}{43.3}$    &   $ \prescript{(62.8)}{}{36.7}$      \\
    \multicolumn{2}{l|}{$\sigma = 1.0$} &  27.7  &  $ \prescript{(50.0)}{}{46.4}$   &   $ \prescript{(50.0)}{}{42.3}$     &    $ \prescript{(50.0)}{}{38.5}$    &   $ \prescript{(50.0)}{}{35.7}$      \\ \hline
    \multicolumn{2}{l|}{\textbf{Mixed Noise}} & \textbf{83.6}  &  $ \prescript{(\textbf{80.1})}{}{\textbf{73.3}}$   &   $ \prescript{(\textbf{80.1})}{}{\textbf{65.4}}$     &    $ \prescript{(\textbf{74.8})}{}{\textbf{56.7}}$    &   $ \prescript{(\textbf{74.8})}{}{\textbf{48.3}}$      \\
    \hline
    \end{tabular}
\end{table*}

\begin{table*}[t]
  \centering
  \caption{Experiments on Flowers102. Mixed noise pre-training enables the model to achieve the best certified accuracy at $\sigma=0.25, 0.5, 0.75$. The clean accuracy is close to that of training with only clean images. 
  }
  \label{tab:flowers}
  \begin{tabular}{cc|c|cccc}
    \hline
    \multicolumn{2}{l|}{\multirow{2}{*}{Method}} & \multirow{2}{*}{\begin{tabular}[c]{@{}c@{}}Clean Acc.\end{tabular}} & \multicolumn{4}{c}{Certified Acc. at $\varepsilon$ (\%)} \\ \cline{4-7} 
    \multicolumn{2}{c|}{}       &                                                                       & 0.25           & 0.5           & 0.75           & 1.0          \\ \hline
    \multicolumn{2}{l|}{Clean} & \textbf{92.2}    & $ \prescript{(14.8)}{}{10.1}$      & $ \prescript{(14.8)}{}{6.8}$          & $ \prescript{(14.8)}{}{4.9}$  &    $ \prescript{(0.8)}{}{0.8}$         \\
    \multicolumn{2}{l|}{$\sigma = 0.25$}  &  31.9  &  $ \prescript{(70.6)}{}{68.7}$   &   $ \prescript{(70.6)}{}{64.8}$     &    $ \prescript{(70.6)}{}{58.6}$    &   $ \prescript{(8.1)}{}{5.5}$      \\
    \multicolumn{2}{l|}{$\sigma = 0.5$} &  26.8  &  $ \prescript{(70.6)}{}{66.6}$   &   $ \prescript{(70.6)}{}{63.1}$     &    $ \prescript{(70.6)}{}{60.1}$    &   $ \prescript{(\textbf{70.6})}{}{\textbf{55.4}}$      \\
    \multicolumn{2}{l|}{$\sigma = 1.0$} &  13.4  &  $ \prescript{(55.7)}{}{54.4}$   &   $ \prescript{(55.7)}{}{52.4}$     &    $ \prescript{(55.7)}{}{51.3}$    &   $ \prescript{(55.7)}{}{50.2}$      \\ \hline
    \multicolumn{2}{l|}{\textbf{Mixed Noise}} & 87.5 &  $ \prescript{(\textbf{78.2})}{}{\textbf{74.4}}$   &   $ \prescript{(\textbf{78.2})}{}{\textbf{70.1}}$     &    $ \prescript{(\textbf{78.2})}{}{\textbf{64.6}}$    &   $ \prescript{(68.0)}{}{53.1}$      \\
    \hline
    \end{tabular}
\end{table*}

\begin{table*}[t]
  \centering
  \caption{Experiments on DTD. Mixed noise pre-training enables the model to achieve the best certified accuracy at $\sigma=0.25, 0.5, 1.0$. The clean accuracy is close to that of training with only clean images.
  }
  \label{tab:dtd}
  \begin{tabular}{cc|c|cccc}
    \hline
    \multicolumn{2}{l|}{\multirow{2}{*}{Method}} & \multirow{2}{*}{\begin{tabular}[c]{@{}c@{}}Clean Acc.\end{tabular}} & \multicolumn{4}{c}{Certified Acc. at $\varepsilon$ (\%)} \\ \cline{4-7} 
    \multicolumn{2}{c|}{}       &                                                                       & 0.25           & 0.5           & 0.75           & 1.0          \\ \hline
    \multicolumn{2}{l|}{Clean} & \textbf{68.6}    & $ \prescript{(24.5)}{}{20.8}$      & $ \prescript{(24.5)}{}{17.2}$          & $ \prescript{(24.5)}{}{14.6}$  &    $ \prescript{(4.2)}{}{4.2}$         \\
    \multicolumn{2}{l|}{$\sigma = 0.25$}  &  31.2  &  $ \prescript{(49.0)}{}{44.8}$   &   $ \prescript{(49.0)}{}{41.7}$     &    $ \prescript{(49.0)}{}{\textbf{39.1}}$    &   $ \prescript{(15.1)}{}{12.0}$      \\
    \multicolumn{2}{l|}{$\sigma = 0.5$} &  10.6  &  $ \prescript{(40.6)}{}{38.0}$   &   $ \prescript{(40.6)}{}{35.9}$     &    $ \prescript{(40.6)}{}{32.8}$    &   $ \prescript{(40.6)}{}{29.7}$      \\
    \multicolumn{2}{l|}{$\sigma = 1.0$} &  5.9  &  $ \prescript{(29.2)}{}{27.1}$   &   $ \prescript{(29.2)}{}{27.1}$     &    $ \prescript{(29.2)}{}{25.5}$    &   $ \prescript{(29.2)}{}{22.9}$      \\ \hline
    \multicolumn{2}{l|}{\textbf{Mixed Noise}} & 66.5 &  $ \prescript{(\textbf{60.9})}{}{\textbf{54.7}}$   &   $ \prescript{(\textbf{60.9})}{}{\textbf{46.4}}$     &    $ \prescript{(60.9)}{}{37.0}$    &   $ \prescript{(50.5)}{}{\textbf{31.3}}$      \\
    \hline
    \end{tabular}
\end{table*}

\section{Experiments on Additional Downstream Tasks}
\label{sec:exp2}
Our mixed noise pre-training model can simultaneously transfer semantic representations and robustness to various downstream tasks.
Here, we present additional experiments on more downstream tasks. We first introduce the datasets and experimental settings. Then, we present the results and analysis.

\subsection{Datasets and Experimental Settings}
we present the experimental results on four additional datasets: Pets, Food101, Flowers102, and DTD. 
Pets is a dataset of 7,349 images of 37 different pet breeds.
Food101 is a dataset of 101,000 images of 101 different food categories. 
Flowers102 is a dataset of 8,189 images of 102 different flower categories.
DTD is a dataset of 5,640 images of 47 different texture categories.
These datasets are primarily used for performing fine-grained classification tasks. DTD is specifically used for texture recognition, while the other three datasets are employed for fine-grained category recognition.

We endeavor to keep the experimental setup consistent with that of CIFAR10 and CIFAR100. One significant difference is that these datasets have a smaller number of images.
Training from scratch on small-scale datasets often leads to poor results.
Therefore, for experiments with only clean images or specific noise, we load the weights pre-trained on ImageNet from the official Pytorch website.
This is a standard configuration in existing randomized smoothing methods. 
For our mixed noise pre-trained model, we perform fine-tuning on downstream tasks using only clean images. Despite fine-tuning solely on clean images, the model is still capable of transferring robustness to downstream tasks.

\subsection{Experiments on Pets}
As shown in Table~\ref{tab:pets}, training with only clean images achieves good clean accuracy (91.3 \%) but poor certified accuracy.
Training with noise at different $\sigma$ can achieve decent certified accuracy, but it comes at the cost of sacrificing clean accuracy.
Pre-training with mixed noise and fine-tuning on clean images from Pets can achieve both good clean accuracy (91.8 \%) and certified accuracy.
The certified accuracy curves are shown in Figure~\ref{fig:pet exp}.
The strong certified accuracy is transferred from the pre-training phase, since the fine-tuning phase only uses clean images.
This demonstrates that our mixed noise pre-training model can simultaneously transfer semantic representations and robustness to downstream tasks.

\subsection{Experiments on Food101}
As shown in Table~\ref{tab:Food101}, training with only clean images achieves good clean accuracy (78.0 \%) but poor certified accuracy.
Training with noise at different $\sigma$ has to sacrifice clean accuracy to achieve good certified accuracy.
Mixed noise pre-training enables the model to achieve the best clean accuracy (83.6 \%) and certified accuracy (at $\sigma = 0.25, 0.5, 0.75, 1.0$).
The certified accuracy curves are shown in Figure~\ref{fig:food101 exp}.

\subsection{Experiments on Flowers102}
As shown in Table~\ref{tab:flowers}, training with only clean images achieves the best clean accuracy (92.2 \%) but poor certified accuracy.
Training with noise at different $\sigma$ has to trade off clean accuracy to achieve good certified accuracy.
Mixed noise pre-training enables the model to achieve the best certified accuracy (at $\sigma = 0.25, 0.5, 0.75$) and decent clean accuracy (87.5 \%).
The certified accuracy curves are shown in Figure~\ref{fig:flowers exp}.

\subsection{Experiments on DTD}
\begin{figure*}
  \centering
  \begin{subfigure}{0.49\linewidth}
    \includegraphics[width=1.0\textwidth]{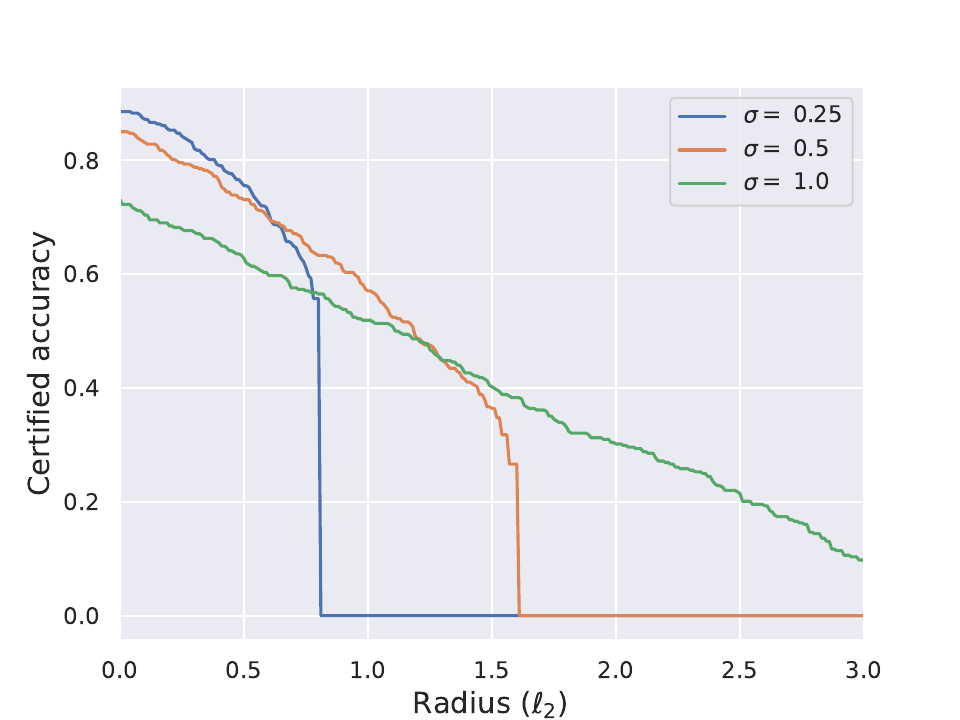}   
    \caption{Pets.}
    \label{fig:pet exp}
  \end{subfigure}
  \hfill
  \begin{subfigure}{0.49\linewidth}
    \includegraphics[width=1.0\textwidth]{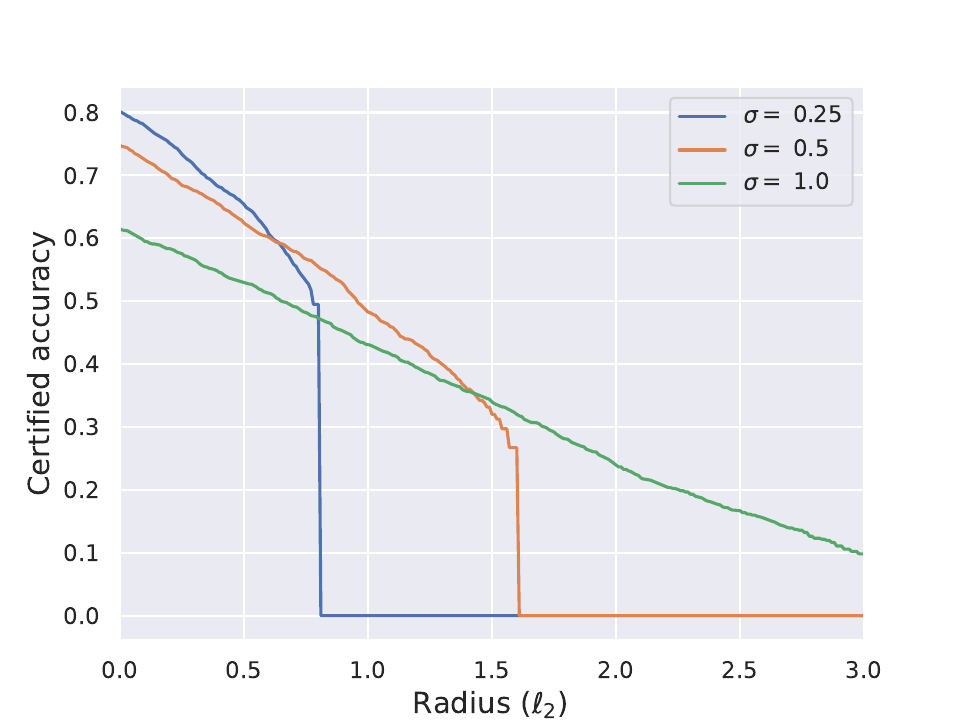}   
    \caption{Food101.}
    \label{fig:food101 exp}
  \end{subfigure}
  \begin{subfigure}{0.49\linewidth}
    \includegraphics[width=1.0\textwidth]{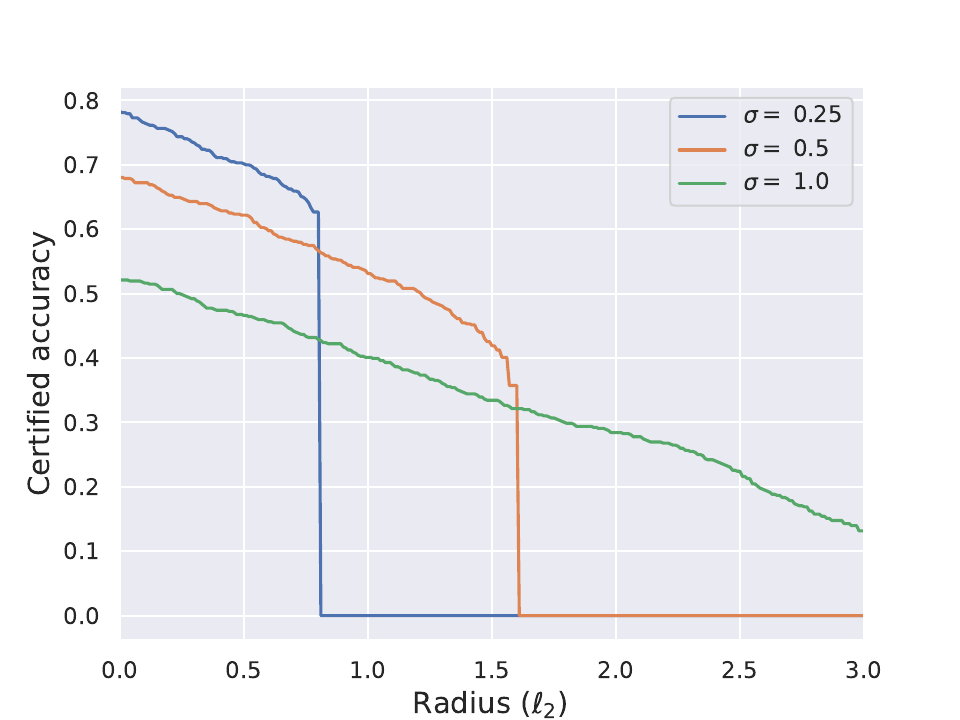}   
    \caption{Flowers102.}
    \label{fig:flowers exp}
  \end{subfigure}
  \hfill
  \begin{subfigure}{0.49\linewidth}
    \includegraphics[width=1.0\textwidth]{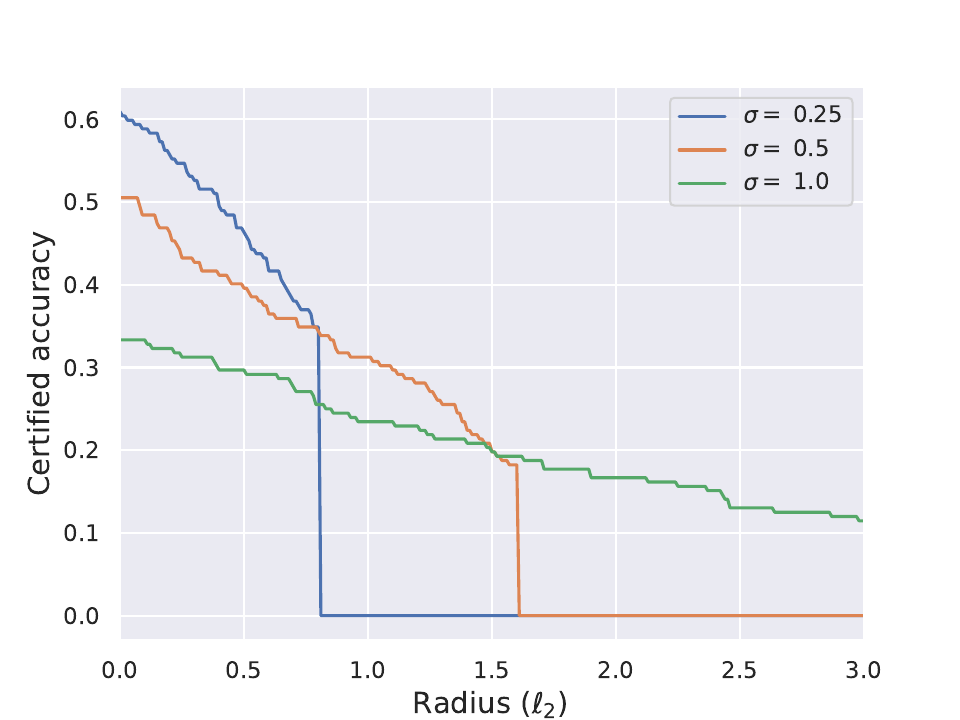}   
    \caption{DTD.}
    \label{fig:dtd exp}
  \end{subfigure}

  \caption{Certified accuracy on (a) Pets, (b) Food101, (c) Flowers102, and (d) DTD.
  We present results with three noise levels $\sigma \in \{0.25, 0.5, 1.0\}$ for each dataset. We require only one model pre-trained on ImageNet with mixed noise. This model can simultaneously transfer semantic representations and robustness to multiple downstream tasks. Also, this single model can handle various levels of noise.}
  \label{fig:downstream1}
\end{figure*}

As shown in Table~\ref{tab:dtd}, training with only clean images achieves the best clean accuracy (68.6 \%) but poor certified accuracy.
Training with noise at different $\sigma$ compromises between clean and certified accuracy.
Mixed noise pre-training enables the model to achieve the best certified accuracy (at $\sigma = 0.25, 0.5, 1.0$) and good clean accuracy (66.5 \%).
The certified accuracy curves are shown in Figure~\ref{fig:dtd exp}.

\end{document}